\documentclass{article}

\usepackage{amsmath,amssymb,amsfonts,amstext}%
\usepackage{graphicx}%
\usepackage{multirow}%
\usepackage{import}
\usepackage{siunitx}

\usepackage[preprint, nonatbib]{neurips_2024}

\usepackage[utf8]{inputenc} 
\usepackage[T1]{fontenc}    
\usepackage{url}            
\usepackage{booktabs}       
\usepackage{nicefrac}       
\usepackage{microtype}      
\usepackage{biblatex} 
\usepackage{array}
\usepackage[caption=false,font=normalsize,labelfont=sf,textfont=sf]{subfig}
\usepackage{textcomp}
\usepackage{stfloats}
\usepackage{url}
\usepackage{amsthm}
\usepackage{lscape}
\usepackage{multirow}
\usepackage{longtable}
\usepackage{caption}
\usepackage[hidelinks]{hyperref}
\usepackage[dvipsnames]{xcolor}
\usepackage{centernot,cancel}
\usepackage{lipsum}
\usepackage{soul,balance}
\usepackage{pifont}

\newcommand{\xmark}{\ding{55}}%

\title{PATH: A Discrete-sequence Dataset for Evaluating Online Unsupervised Anomaly Detection Approaches for Multivariate Time Series}

\author{%
  Lucas Correia\\
  Leiden University \\
  Leiden \\
  The Netherlands \\
  \texttt{l.ferreira.correia@liacs.leidenuniv.nl} \\
  \And
  Jan-Christoph Goos \\
  Mercedes-Benz AG\\
  Stuttgart\\
  Germany \\
  \AND
  Thomas Bäck \\
  Leiden University \\
  Leiden \\
  The Netherlands \\
  \And
  Anna V. Kononova \\
  Leiden University \\
  Leiden \\
  The Netherlands \\
\texttt{a.kononova@liacs.leidenuniv.nl} \\
}

\newcommand{\lc}[1]{\textcolor{black}{#1}}

\begin{document}

\maketitle

\begin{abstract}
Benchmarking anomaly detection approaches for multivariate time series is \lc{a} challenging task due to \lc{a} lack of high-quality datasets. 
Current publicly available datasets are too small, not diverse and feature trivial anomalies, which hinders measurable progress in this research area.
We propose a solution: a diverse, extensive, and non-trivial dataset generated via state-of-the-art simulation tools that reflects realistic behaviour of an automotive powertrain, including its multivariate, dynamic and variable-state properties. 
\lc{Additionally, our dataset represents a discrete-sequence problem, which remains unaddressed by previously-proposed solutions in literature.}
To cater for both unsupervised and semi-supervised anomaly detection settings, as well as time series generation and forecasting, we make different versions of the dataset available, where training and test subsets are offered in contaminated and clean versions, depending on the task.
We also provide baseline results from a selection of approaches based on deterministic and variational autoencoders, as well as a non-parametric approach.
As expected, the baseline experimentation shows that the approaches trained on the semi-supervised version of the dataset outperform their unsupervised counterparts, highlighting a need for approaches more robust to contaminated training data.
\lc{Furthermore, results show that the threshold used can have a large influence on detection performance, hence more work needs to be invested in methods to find a suitable threshold without the need for labelled data.}
\end{abstract}

\section{Introduction}\label{sec:introduction}
As the digitisation of industrial processes progresses, more and more data is recorded. 
Ensuring this data is representative of the process is important, as downstream tasks like modelling or optimisation can be negatively impacted by incomplete or contaminated data.
For tasks that require system behaviour modelling, data deviating from the norm is hence undesired, and we speak of \emph{anomalous} behaviour.
Recorded data manifests itself in many forms depending on the application and domain, one form being time series. 
Examples of real-world time series applications are diverse, ranging from cardiology~\cite{moody_impact_2001} and server metrics monitoring~\cite{su_robust_2019} to water systems~\cite{mathur_swat_2016, ahmed_wadi_2017, zhang_anomaly_2023} and unmanned aerial vehicles~\cite{zhong_unmanned_2022}.
Note that, we use \emph{time series} and \emph{sequences} synonymously throughout this paper.

\emph{Time series} are signals that represent a property or feature of a dynamic system as a function of time, usually sampled at a fixed rate. 
An arbitrary time series $\mathcal{S}$ can be univariate, i.e.\ $\mathcal{S} \in \mathbb{R}^{T}$, or multivariate, i.e.\ $\mathcal{S} \in \mathbb{R}^{T \times d}$, where $T$ refers to the number of discrete time steps and $d$ to the number of features in the time series.
More specifically, univariate time series solely possess a temporal correlation, i.e.\ along the time axis, whereas multivariate time series can also contain correlation along the feature axis.

Detecting anomalous behaviour in time series is referred to \emph{time series anomaly detection}, which can be split into two main areas: \emph{continuous- and discrete-sequence}~\cite{correia_online_2024}, where the former is \lc{the only type addressed} in public datasets.
Continuous-sequence problems are defined as detecting anomalies in a process that runs for a continuous time period without breaks. 
Typically, the test subset $\mathcal{D}^\text{test}$ in the dataset $\mathcal{D}$ in a continuous-sequence problem consists of a singular multivariate time series composed of multiple nominal and anomalous sub-sequences, i.e.\ $\mathcal{D}^\text{test} = \{\mathcal{S}_1\}$.
In this work, we use \emph{nominal} as a synonym for \emph{normal} or \emph{anomaly-free} to avoid confusion with Gaussian distributions.
Discrete-sequence anomaly detection, in contrast, is defined as detecting anomalies in $N$ chunks of processes that happen independently of each other.
\lc{Discrete-sequence problems include, for example, } automotive test benches, where several tests may occur sequentially but are not temporally contiguous and hence provide a time series for each test, i.e.\ $\mathcal{D}^\text{test} = \{\mathcal{S}_1, ..., \mathcal{S}_n, ..., \mathcal{S}_{N}\}$ and $N>1$.
Here, the testees, i.e.\ the test subjects, are not monitored over a continuous period of time, but are instead monitored solely during each process chunk.
Automotive testing is not the only use for discrete-sequence anomaly detection, however.
Another discrete-sequence problem could also include the analysis of the flight behaviour of an aeroplane, where the time while it is docked is irrelevant and may not be recorded.
Therefore, datasets for discrete-sequence anomaly detection consist of several nominal and anomalous time series, where a given anomalous time series may be entirely anomalous or only partly.
Depending on the system, the time series data may also feature \emph{variable states}, meaning the recorded signals appear slightly different if certain external conditions change but are still considered nominal.
One example of a variable-state system is a battery, where the voltage response to current changes depending on states like the battery temperature and the battery state of charge (SoC).
A problem involving such a system requires the distinction between behaviour changes due to different states and behaviour changes due to an anomaly, further complicating detection.

In addition to that, detecting anomalous behaviour in a timely manner is also advantageous because the source of anomalous behaviour may bring about damage to said system.
Problems where the detection delay plays a role \lc{require evaluation of the time series before it ends} and are referred to as \emph{online} time series anomaly detection.

Analogous to types of learning, there is supervised, semi-supervised, and unsupervised anomaly detection.
\emph{Supervised} anomaly detection is essentially imbalanced binary time series classification and is only rarely found in the literature. 
This is most likely because, in real-world problems, possible anomaly types \lc{and how they manifest themselves in the data} are rarely known a priori.
Moreover, labelling data is expensive, which is why unsupervised and semi-supervised anomaly detection are more prevalent in literature and relevant to the real world.
\emph{Unsupervised} anomaly detection is independent of any labels, i.e.\ any available data for model training contains both anomalous and nominal time series, and it is not known which is which~\cite{chandola_anomaly_2012}.
In contrast, \emph{semi-supervised} anomaly detection can be considered a more relaxed setting, where anomalous time series are absent from the training subset~\cite{chandola_anomaly_2012}.
In the real world, semi-supervised problems still require some labelling to ensure an entirely nominal training subset, which is not always given.
\lc{Some literature diverges from this taxonomy, agreeing with the supervised and semi-supervised definitions but defining unsupervised anomaly detection differently.
According to Schmidl et al.~\cite{schmidlAnomalyDetectionTime2022}, unsupervised anomaly detection involves detecting anomalous behaviour without a training procedure.
This definition implies that learning nominal behaviour from mostly-nominal data is not possible, which we \lc{challenge} in this paper.}


Our contribution is a non-trivial, and high-quality \lc{discrete-sequence} dataset consisting of multivariate time series for online anomaly detection, named the Powertrain Anomaly Time series bencHmark (PATH) dataset.
While primarily aimed at unsupervised anomaly detection, we provide versions for semi-supervised anomaly detection and time series generation and forecasting as well. 
Despite the data being generated using simulation, the electric vehicle simulation model is strongly motivated by the real world and is therefore complex and variable-state.

This paper is structured as follows.
First, we introduce the related work in the area of benchmarking time series anomaly detection approaches.
It includes discussion on the datasets used to evaluate time series anomaly detection approaches in the past, and a summary of the work dedicated to outlining the status quo in benchmarking time series anomaly detection approaches.
Then, we introduce the PATH dataset in detail, outlining the generation process and a few benchmarking considerations.
Following that, we provide some baseline results for a selection of deep learning-based models, as well as a non-parametric approach.
Finally, we conclude our work and outline an outlook on future work.
The source code corresponding to this paper and the simulation model can be found on \lc{\href{https://github.com/lcs-crr/PATH}{Github}}, and the dataset can be downloaded from \lc{\href{https://doi.org/10.5281/zenodo.13255120}{Zenodo}}~\cite{data_set}.

\section{Related Work} \label{sec:related_work}
\subsection{Publicly Available Datasets}
Over the last few years, five benchmark datasets have emerged as by far the most popular, with at least one of them being cited in the vast majority of publications on multivariate time series anomaly detection. 
A summary of the properties of these datasets is shown in Table~\ref{tab:public_data_set_properties}.

\begin{table}[t!]
    \centering
    \caption{Key properties of the most popular datasets, where $d_\mathcal{D}$ refers to the number of features of the time series signals in the dataset, \lc{$\text{DS}$  to whether it is a discrete-sequence dataset, $\sum \vert \mathcal{S}_n\vert$ to the number of test time steps and $\%A$ to the number of anomalous time steps in relation to all test time steps. GutenTAG varies in $d_\mathcal{D}$ depending on test time series. The number of time steps given for the test subset in PATH is the average across all folds.}}
    \label{tab:public_data_set_properties}
    \begin{tabular}{lcccc}
    Name                                                                         & $d_\mathcal{D}$ & $\text{DS}$       & $\displaystyle\sum^N\vert\mathcal{S}_n\vert$ & $\%A$                \\ \hline\hline
    \href{https://itrust.sutd.edu.sg/itrust-labs_data sets/data set_info/}{SWaT} & $51$            & \xmark     & $449,919$                                    & \qty{12}{\percent}   \\
    \href{https://itrust.sutd.edu.sg/itrust-labs_data sets/data set_info/}{WADI} & $127$           & \xmark     & $17,287$                                     & \qty{6}{\percent}    \\
    \href{https://github.com/khundman/telemanom}{SMAP}                           & $25$            & \xmark     & $427,617$                                    & \qty{13}{\percent}   \\
    \href{https://github.com/khundman/telemanom}{MSL}                            & $55$            & \xmark     & $73,729$                                     & \qty{11}{\percent}   \\
    \href{https://github.com/NetManAIOps/OmniAnomaly}{SMD}                       & $38$            & \xmark     & $708,420$                                    & \qty{4}{\percent}    \\ 
    \href{https://github.com/TimeEval/GutenTAG}{GutenTAG}                        & $2-20$          & \xmark     & $240,000$                                    & \qty{1}{\percent}    \\     
    \href{https://github.com/2er0/mTADS}{mTADS}                                  & $4$             & \xmark     & $2,396,000$                                  & $<$\qty{1}{\percent} \\     
    \textbf{Ours}                                                                & $16$            & \checkmark & $14,341,432$                                          & \qty{7}{\percent}                  \\ \hline
    \end{tabular}
\end{table}

The MSL~\cite{hundman_detecting_2018}, SMAP~\cite{hundman_detecting_2018}, and SMD~\cite{su_robust_2019} have already been thoroughly analysed by Wu and Keogh~\cite{wu_current_2021}, who point out several \emph{issues} with the datasets. 
The first issue observed in the datasets is triviality, defined by being solvable using so-called \emph{one-line code}, such as the moving standard deviation over a subset of the dataset features. 
Moreover, all of them suffer from what Wu and Keogh~\cite{wu_current_2021} have called unrealistic anomaly density, meaning that they have sub-sequences with a very high anomaly share and hence do not match the assumption that anomalies are rare events.
In addition to that, Wu and Keogh~\cite{wu_current_2021} suspect possible mislabelling present in the MSL dataset. 
They base their suspicion on the fact that the dataset contains sub-sequences with static behaviour in an evidently dynamic channel, which is labelled as nominal.
\lc{Additionally, while the SMAP and MSL datasets are technically multivariate, the channels are not synchronised with each other and hence each channel needs to be modelled on its own.}

Many of the issues pointed out by Wu and Keogh~\cite{wu_current_2021} can also be extended to the SWaT and WADI datasets, as thoroughly discussed by Wagner et al.~\cite{wagnerTimeSeADBenchmarkingDeep2023} and Correia et al.~\cite{correia_online_2024}. 
\lc{Both datasets have multiple features that are constant throughout the training and testing subsets, while WADI has some features with missing values.
This leads to ambiguity when benchmarking, as some may choose to omit these features while others may not.
Furtermore, $65\%$ of the anomalous time steps in the SWaT dataset can be detected by simply inspecting one feature.}

\lc{Additionally, Wenig et al.~\cite{wenigAnomalyDetectorsMultivariate2024} point out that the SWaT, WADI and SMD datasets feature mostly univariate anomalies, i.e.\ anomalous behaviour manifests itself in a single channel.
They found that, in these datasets, univariate approaches mostly outperform multivariate approaches due to the overwhelming presence of such univariate anomalies.
However, once multivariate anomalies are injected, multivariate approaches outperform their univariate counterparts.}

\lc{ Wu and Keogh~\cite{wu_current_2021} have caused a shift towards more transparency when benchmarking time series anomaly detection methods.
Since then, some contributions have been made to address the aforementioned issues, also shown in Table~\ref{tab:public_data_set_properties}.}

\lc{As part of the TimeEval framework, Wenig et al.~\cite{wenigTimeEvalBenchmarkingToolkit2022} propose the Good Time Series Anomaly Generator (GutenTAG), a tool for time series dataset generation, which can generate nominal and anomalous multivariate time series.
It works by combining base oscillations, like sine or ECG-like waves, and injecting different types of pre-defined anomalies.
However, GutenTAG only represents the tool to create a dataset, not the dataset itself. 
Further fragmentation of the research field will occur if no dataset resulting from GutenTAG is agreed upon, though GutenTAG-based data used in a benchmarking paper by the same authors~\cite{schmidlAnomalyDetectionTime2022} may serve as a reference.
While the time series data generated by GutenTAG may be complex due to the different combinations of base oscillations, it still lacks the relationship to the real world.
Arguably, the main purpose of research on time series anomaly detection is to solve real-world problems, and hence any evaluation should also consider real-world or real-world-inspired data.}

\lc{mTADS~\cite{baumgartnerMTADSMultivariateTime2023} is a collection of two types of datasets, one completely synthetic generated with GutenTAG and another based on simulation of the Lotka-Volterra equations which represent the relationship between one predator, two prey populations and another population that's both predator and prey.
One interesting aspect of this dataset is that training data is offered both with and without anomalous behaviour, allowing for semi- and unsupervised anomaly detection. 
Despite being real world-inspired the two equations are still fairly simple, yielding time series data with only four features.}


Certain real-world applications like automotive testing present \emph{complexities} previously unseen in public datasets. 
As outlined by Correia et al.~\cite{correiaTeVAEVariationalAutoencoder2025}, such applications comprise much more diverse discrete-sequence datasets, owed to the presence of both highly dynamic and mostly static features, as well as variable states. 
The presence of variable states leads to features exhibiting a different pattern depending on the time it is observed. 
In the context of automotive testing, an example of such a feature would be the state of charge of a battery, which discharges with time and hence shows different behaviour for the same test done twice in a row.

Hence, we construct a new high-quality dataset that features non-trivial anomalies \lc{caused by pre-simulation model changes} and that \lc{mostly} reflects real-world complexity. 
\lc{Additionally, as a discrete-sequence dataset, it represents a contribution to an underrepresented problem type that often occurs in the real world. 
Since the dataset is generated using simulation, it also strikes the balance between real-world relevance and control over anomalous behaviour~\cite{baumgartnerMTADSMultivariateTime2023}.}

\subsection{Doubts Regarding Applicability of Deep Learning}
Recently, there has been growing doubt on whether deep learning (DL) algorithms are definitively the better choice for time series anomaly detection. 
For the purpose of this publication, classical methods refer to all approaches not based on DL, including non-parametric and statistical approaches, as well as simpler machine learning methods like clustering.

Wu and Keogh~\cite{wu_current_2021} claim that the superiority of DL in anomaly detection is assumed to be a given, despite a lack of clear evidence for the need for DL. 
They stress that existing classical methods should be considered, given their generally simpler and faster nature.

To investigate the comparative performance of classical methods and DL-based methods, Audibert et al.~\cite{audibert_deep_2022} analyse a variety of different models on five of the most popular benchmark datasets, shown in Table~\ref{tab:public_data_set_properties}. 
They conclude that, across the datasets considered, there is no algorithm that dominates all the other ones, arguing that there is no reason to omit classical methods from benchmarking. 

Rewicki et al.~\cite{rewicki_is_2023} also conduct a comparative study of classical and DL-based methods, though on the UCR Anomaly Archive benchmark proposed by Wu and Keogh~\cite{wu_current_2021}, which exclusively contains univariate time series and therefore lacks correlations between channels present in multivariate time series. 
They conclude that classical methods perform better than their DL counterparts, although this is to be expected given the simpler, univariate nature of the dataset.

While the findings and doubts of the above-mentioned are valid, they are limited by the lack of large, high-quality multivariate datasets. 
In this paper, we purposefully include results from a state-of-the-art classical method to find out whether doubts on DL are still justified for extensive and complex real-world\lc{-inspired} datasets.
See Section~\ref{sec:results} for results and discussion.

\section{Proposed Dataset} \label{sec:dataset}
\subsection{Simulation Model}
\begin{figure}[t!]
    \centering
    \includegraphics[width=0.80\textwidth]{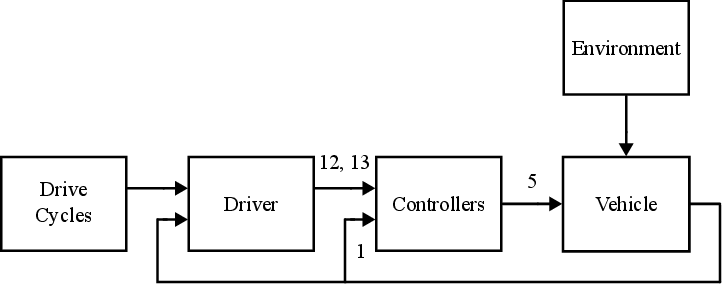}
    \caption{Simplified schematic of the FEV model used for the generation of the PATH dataset. Numbers represent the indices of signal flow, reference is shown in Table~\ref{tab:signals}.}
    \label{fig:sim_model}
\end{figure}

To create an extensive and diverse dataset, we propose to use a physically inspired model, from which we can generate data using simulation. 
MathWorks offers reference models for a variety of dynamic systems, one of which is the full electric vehicle (FEV) model\footnote{https://de.mathworks.com/help/autoblks/ug/explore-the-electric-vehicle-reference-application.html, last accessed: 24.03.2025} from the powertrain blockset in Simulink. 
This choice is based on our familiarity with the domain, as generating data blindly without any background may lead to systematic errors.
The FEV model offered by MathWorks consists of six main subsystems: the drive cycle block, the driver block, the environment block, the controllers block and the vehicle block. 
The topology of the FEV model is illustrated in Figure~\ref{fig:sim_model}.

To represent system behaviour, $d_\mathcal{D}=16$ signals are chosen to be logged during simulation and are summarised in Table~\ref{tab:signals}. 
We chose these signals based on domain knowledge, with the goal of picking the features that are most representative of powertrain behaviour.
\begin{table}[t!]
\centering
\caption{Signals included in the PATH dataset, along with their physical units and persistent indices.}\label{tab:signals}
\begin{tabular}{clc}
Index   & Signal Name         & Unit  \\ \hline \hline
1       & Motor Speed         & \si{\radian\per\second} \\
2       & Motor Torque        & \si{\newton\metre}    \\
3       & Axle Torque Front   & \si{\newton\metre}    \\
4       & Axle Torque Rear    & \si{\newton\metre}    \\
5       & Battery SoC         & \si{\percent}    \\
6       & Battery Current     & \si{\ampere}     \\
7       & Battery Power       & \si{\watt}     \\
8       & Axle Force Front    & \si{\newton}     \\
9       & Axle Force Rear     & \si{\newton}     \\
10      & Axle Speed Front    & \si{\radian\per\second} \\
11      & Axle Speed Rear     & \si{\radian\per\second} \\
12      & Accelerator Pedal   & -     \\
13      & Brake Pedal         & -     \\
14      & Battery Temperature & \si{\degreeCelsius}    \\
15      & Cooling Pump Power  & \si{\watt}     \\
16      & Refrigerator Power  & \si{\watt}     \\\hline
\end{tabular}
\end{table}

The drive cycle block of the FEV model defines the target vehicle speed and features a series of real-world drive cycles, i.e.\ profiles depicting the target vehicle speed as a function of time. 
From the list of speed profiles available in the original FEV model, a subset is eliminated due to their unrealistic nature, e.g.\ high linearity or duplicity, as many cycles are present as sub-sequences in others. 
Our analysis shows that, for example, the presence of the FTP72 drive cycle within FTP75 or the presence of the LA92Short drive cycle within LA92. 
In addition to that, drive cycles aimed at heavy vehicles, like trucks or buses, are also eliminated. 
The resulting subset of drive cycles chosen for simulation contains $33$ different speed profiles of varying length, shown in Table~\ref{tab:drive_cycles}, along with their lengths in seconds. 
Some drive cycles may be designed for specific types of powertrains such as diesel ones, but given that they only represent vehicle speed profiles, there is little reason why they cannot be driven by a vehicle with an electric powertrain, like the one modelled in this case. 
\begin{table}[t!]
\centering
\caption{Drive cycles used for the PATH dataset generation, along with their respective lengths in seconds.}\label{tab:drive_cycles}
\begin{tabular}{lc}
Drive Cycle                                                 & Length {[}s{]} \\ \hline \hline
FTP75                                                       & 2474           \\
US06                                                        & 600            \\
SC03                                                        & 600            \\
HWFET                                                       & 765            \\
NYCC                                                        & 598            \\
HUDDS                                                       & 1060           \\
LA92                                                        & 1435           \\
IM240                                                       & 240            \\
UDDS                                                        & 1369           \\
WLTP Class 1                                                & 1022           \\
WLTP Class 2                                                & 1477           \\
WLTP Class 3                                                & 1800           \\
Artemis Urban                                               & 993            \\
Artemis Rural Road                                          & 1082           \\
Artemis Motorway 130 kmph                                   & 1068           \\
Artemis Motorway 150 kmph                                   & 1068           \\
JC08                                                        & 1204           \\
JC08 Hot                                                    & 1376           \\
World Harmonized Vehicle Cycle (WHVC)                       & 900            \\
Braunschweig City Driving Cycle                             & 1740           \\
RTS 95                                                      & 886            \\
ETC FIGE Version 4                                          & 1800           \\
CUEDC Petrol cycle                                          & 499            \\
CUEDC SPC240 cycle                                          & 240            \\
CUEDC diesel cycle - MC                                     & 1723           \\
CUEDC diesel cycle - NA                                     & 1795           \\
CUEDC diesel cycle - NB                                     & 1706           \\
CUEDC diesel cycle - ME                                     & 1678           \\
CUEDC diesel cycle - NC                                     & 1797           \\
CUEDC diesel cycle - NCH                                    & 1676           \\
China Light-Duty Vehicle Test Cycle for Passenger Cars      & 1800           \\
China Light-Duty Vehicle Test Cycle for Commercial Vehicles & 1800           \\
China Worldwide Transient Vehicle Cycle                     & 1799           \\\hline
\end{tabular}
\end{table}

The driver block of the FEV model regulates the dynamic system to maintain the target speed. 
Its inputs are the target vehicle speed and the actual vehicle speed, and its outputs are the acceleration and deceleration control commands, index $12$ and $13$ in Table~\ref{tab:signals}, respectively, which are fed into the controllers block of the FEV model. 
This block takes said accelerator and brake pedal commands stemming as well as vehicle states like actual vehicle speed, electric motor speed and battery signals to calculate request signals for the powertrain, like the required electric motor torque and the brake signal, as well as battery management system signals like the battery SoC, index $5$ in Table~\ref{tab:signals}. 
Electric vehicles are capable of regenerative braking, meaning, the electric motor is used to decelerate the vehicle by acting as a generator, thereby charging the battery if it is not already fully charged. 
\begin{figure}[t!]
    \centering
    \includegraphics[width=0.80\textwidth]{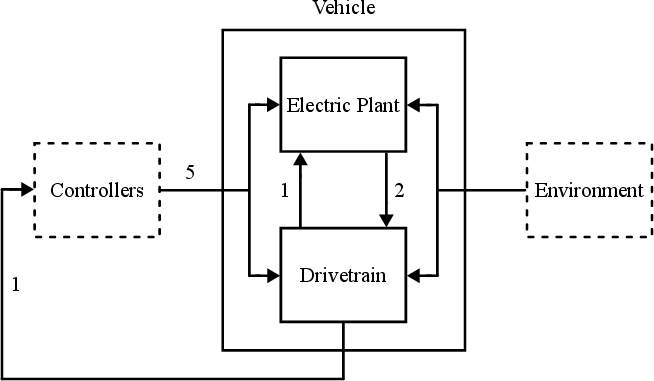}
    \caption{A more detailed schematic of the vehicle model depicted in Figure~\ref{fig:sim_model}. Numbers represent the indices of signal flow, reference is shown in Table~\ref{tab:signals}. Output signals of the vehicle model, which are not fed back into other subsystems, are not shown, for simplicity.}
    \label{fig:veh_model}
\end{figure}

Following is the vehicle block of the FEV model, which outputs how the vehicle reacts to any inputs and contains the electric plant subsystem and the drivetrain subsystem.
Both take inputs from the controllers block, including the battery SoC, and the environment block, as well as from each other, as shown in Figure~\ref{fig:veh_model}.
The electric plant subsystem outputs the electric motor torque, the battery current and power, the battery temperature and the cooling pump and refrigerator powers, which correspond to indices $2$, $6$, $7$, $14$, $15$ and $16$ in Table~\ref{tab:signals}, respectively.
The electric motor torque is input into the drivetrain subsystem, which in turn outputs the electric motor speed, and front and rear axle torques, forces and speeds, corresponding to indices $1$, $3$, $4$, $8$, $9$, $10$, $11$ in Table~\ref{tab:signals}, respectively.
The motor speed is also fed back into the electric plant model, completing the control loop.
Both subsystems also contain further subsystems within them which uncover the causal relationships between their respective signals, but diving as deep as the lowest abstraction level of the model is outside the scope of this paper.
Readers interested in more detail can refer to the Simulink model available in the provided repository.
By default, the battery model features a static temperature model, however, to increase system complexity, a dynamic temperature model is added to the FEV model. 
The model used is the EV Battery Cooling System\footnote{https://de.mathworks.com/help/hydro/ug/ev-battery-cooling.html, last accessed: 24.03.2025}, also from MathWorks.

The environment block of the FEV model dictates environmental conditions that affect the longitudinal dynamics of the FEV model. 
Parameters like atmospheric pressure, wind speed, road grade and coefficient of friction can be set within this subsystem. 

By default, the signals are logged at a sampling frequency of \qty{10}{\hertz} and the solver used is the differential algebraic equations' solver for Simscape (daessc). 
Physical simulations may encounter numerical under- and overflow, which slow down simulations drastically. 
To avoid this, a timeout counter of one hour is set in place to skip the current simulation if triggered. 
Simulation time depends on the length of the drive cycle, however, for the computer hardware used simulations never take longer than 20 minutes, and hence one hour is considered sufficient for problem-free simulations.

\subsection{Dataset Generation}
\begin{figure}[t!]
    \centering
    \includegraphics[width=0.98\textwidth]{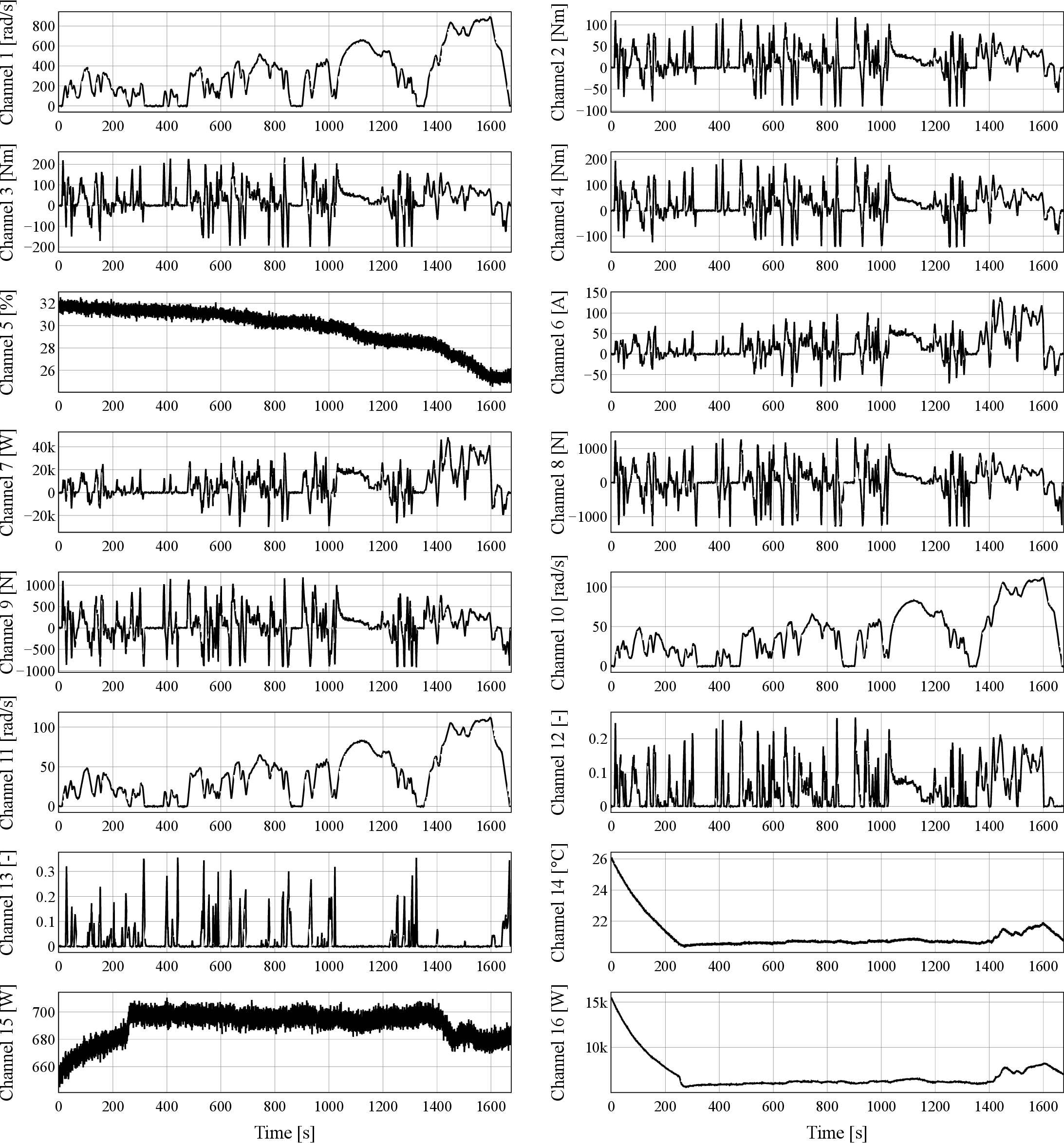}
    \caption{Sample plot of a \emph{nominal} sequence with added noise and undergone trimming. The channel legend can be found in Table~\ref{tab:signals}.}
    \label{fig:data_plot}
\end{figure}

To generate a dataset that is not only extensive but also diverse, $100$ simulations have been undertaken for each of the $33$ drive cycles, each with random initial battery temperatures and battery SoCs. 
At this stage, all model properties have been left as default, and hence all simulation results have been considered \emph{nominal}.
For each simulation, the two states (battery temperatures and battery SoCs) have been sampled from uniform distributions $\mathcal{U}(\qty{10}{\degreeCelsius}, \qty{30}{\degreeCelsius})$ and $\mathcal{U}(\qty{10}{\percent}, \qty{100}{\percent})$, respectively, to ensure no bias is introduced. 
Sampling from uniform distributions also reduces the effectiveness of simple threshold and control chart methods because the battery temperature and state of charge, but also, by extension, other channels, exhibit nominal but high deviation from the average behaviour.
As a precautionary measure, drive cycles with a minimum SoC value lower than \qty{5}{\percent} have been removed, as very low values have been observed to result in abnormal behaviour.
After simulation, $N_\text{n}=3273$ highly diverse and unique nominal time series have been collected.
For illustration purposes, a nominal time series is plotted in Figure~\ref{fig:data_plot}.

For the generation of \emph{anomalous} time series, six types of anomalies have been considered. 
Some types can occur as both sub-sequence anomalies and sequence anomalies~\cite{correia_online_2024}, while others only in sequence anomaly form, due to simulation model limitations. 
To better distinguish the two anomaly forms, we refer to sequence anomalies as \emph{whole-sequence} anomalies henceforth.
The distribution of sub-sequence and whole-sequence anomalies across the different anomaly types is shown in Table~\ref{tab:anomaly_types}.
Anomalies are caused by changing certain model properties \emph{prior} to simulation, ensuring that any observed anomalous behaviour results from simulation rather than manual tampering of the data, like in the UCR dataset~\cite{dau_ucr_2019}, which eliminates any bias. 
We ran all simulations with a fixed seed of $1$.
\begin{table}[t!]
\centering
\caption{Number of sub-sequence anomalies $N_\text{ss}$ and number of 
whole-sequence anomalies $N_\text{ws}$ by anomaly type.}
\label{tab:anomaly_types}
\begin{tabular}{lcc}
Anomaly Types                  & $N_\text{ss}$ & $N_\text{ws}$ \\ \hline\hline
Regenerative Braking Off       & 33 & 32 \\
Increased Headwind             & 33 & 31 \\
Reduced Pump Displacement      & 1  & 23 \\
Reduced Motor Torque Request   & 32 & 33 \\
Increased Wheel Diameter       & 0  & 33 \\
Increased Driver Reaction Time & 0  & 33 \\\hline
\end{tabular}
\end{table}

\begin{figure}[t!]
    \centering
    \includegraphics[width=0.98\textwidth]{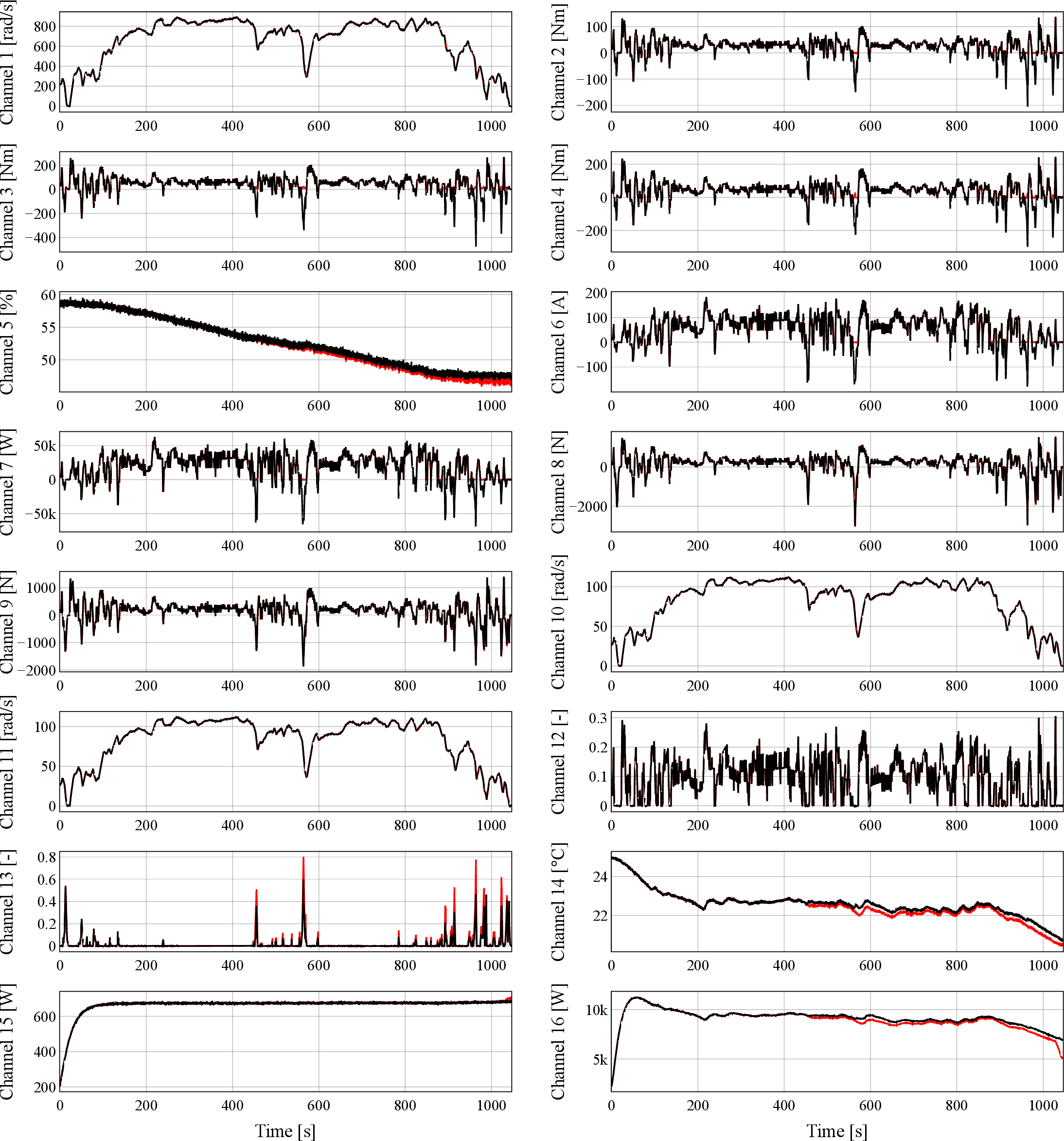}
    \caption{Plot of an \emph{anomalous} sequence without regenerative braking (in red) and its control counterpart (in black), both with added noise and undergone trimming. The anomalous sub-sequence starts after \qty{384.6}{\second}. The channel legend can be found in Table~\ref{tab:signals}.}
    \label{fig:anomaly_plot_brakeregen}
\end{figure}

For the first kind of anomaly, we turn the regenerative braking off, which leads to visibly different motor and axle torques, as well as battery current and power, as these can no longer assume negative values.
When regenerative braking is off, the battery SoC now has an exclusively negative gradient as it is no longer recharged via regenerative braking, and hence it decreases at a faster rate. 
The brake pedal is also used more to compensate for the missing braking motor torque.
For each of the cycles, this anomaly type is simulated in two different ways: without regenerative braking from the beginning and from a random point in time within the drive cycle.
This random point in time is sampled from a uniform distribution $\mathcal{U}(0.2T, 0.8T)$, where $T$ denotes the temporal length of the drive cycle in question, see Table~\ref{tab:drive_cycles}.
This statistical distribution is used for all sub-sequence anomaly types.
One of the anomalous time series for the CADC130 drive cycle and its control counterpart are plotted in Figure~\ref{fig:anomaly_plot_brakeregen}. 

\begin{figure}[t!]
    \centering
    \includegraphics[width=0.98\textwidth]{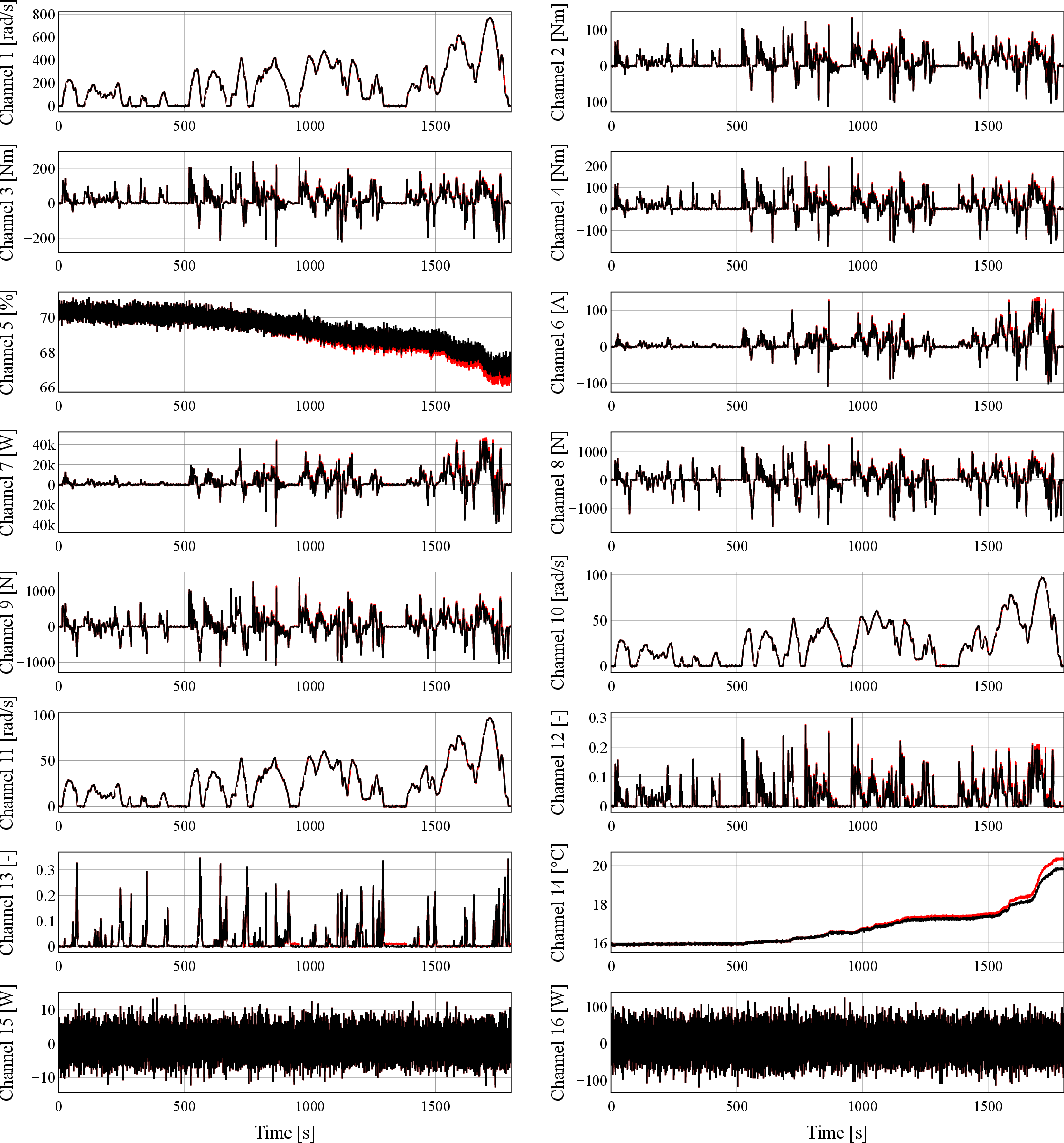}
    \caption{Plot of an \emph{anomalous} sequence with an added headwind (in red) and its control counterpart (in black), both with added noise and undergone trimming. The anomalous sub-sequence starts after \qty{738.0}{\second}. The channel legend can be found in Table~\ref{tab:signals}.}
    \label{fig:anomaly_plot_extwind}
\end{figure}

In the case of the next anomaly type, we introduce a headwind of \qty{5}{\metre\per\second} to the model. 
This headwind acts as a force on the frontal area of the vehicle and needs to be overcome to maintain the target vehicle speed by using the accelerator pedal more than the norm, which leads to higher motor and axle torques and therefore axle forces.
The higher motor torque requires a higher battery current and power, which also causes accelerated discharging.
Like previously, this anomaly type is simulated for each drive cycle, both from the beginning and from a random point in time within the cycle.
One of the anomalous time series for the CLTCPassenger drive cycle and its control counterpart are plotted in Figure~\ref{fig:anomaly_plot_extwind}. 

\begin{figure}[t!]
    \centering
    \includegraphics[width=0.98\textwidth]{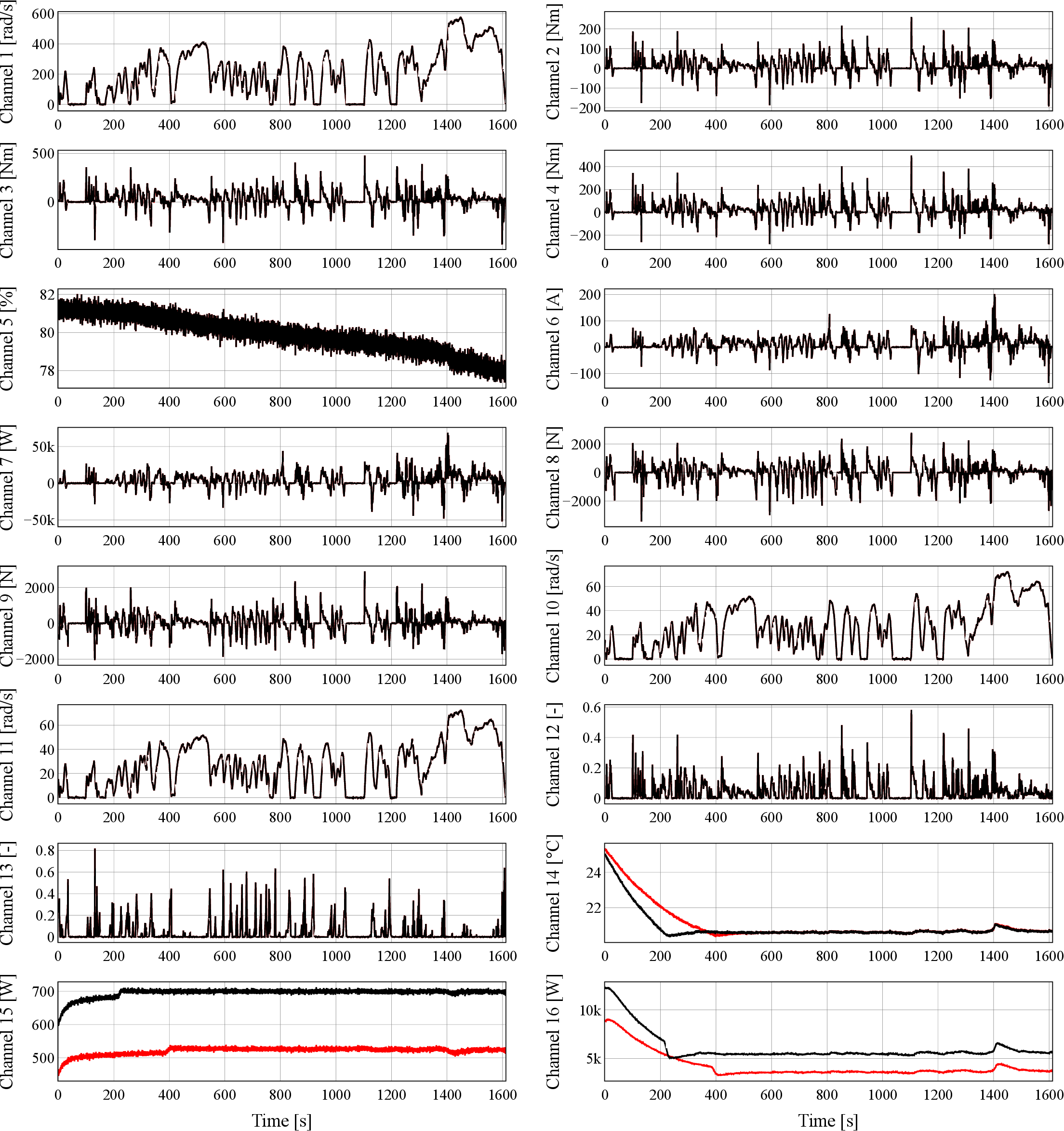}
    \caption{Plot of an \emph{anomalous} sequence with a reduced cooling pump displacement (in red) and its control counterpart (in black), both with added noise and undergone trimming. It is a whole-sequence anomaly, and hence the anomalous behaviour starts from the first time step. The channel legend can be found in Table~\ref{tab:signals}.}
    \label{fig:anomaly_plot_pumpspeed}
\end{figure}

Following that, we reduce the displacement of the cooling pump by \qty{10}{\percent} to simulate another anomaly type. 
Evidently, this change leads to a higher battery temperature as the cooling capacity is reduced. 
This reduction is also visible in the pump power. 
Like with the previous two anomaly types, this anomaly type can start from the beginning and from a random point in time within the cycle.
One of the anomalous time series for the CUEDCDieselME drive cycle and its control counterpart are plotted in Figure~\ref{fig:anomaly_plot_pumpspeed}. 

\begin{figure}[t!]
    \centering
    \includegraphics[width=0.98\textwidth]{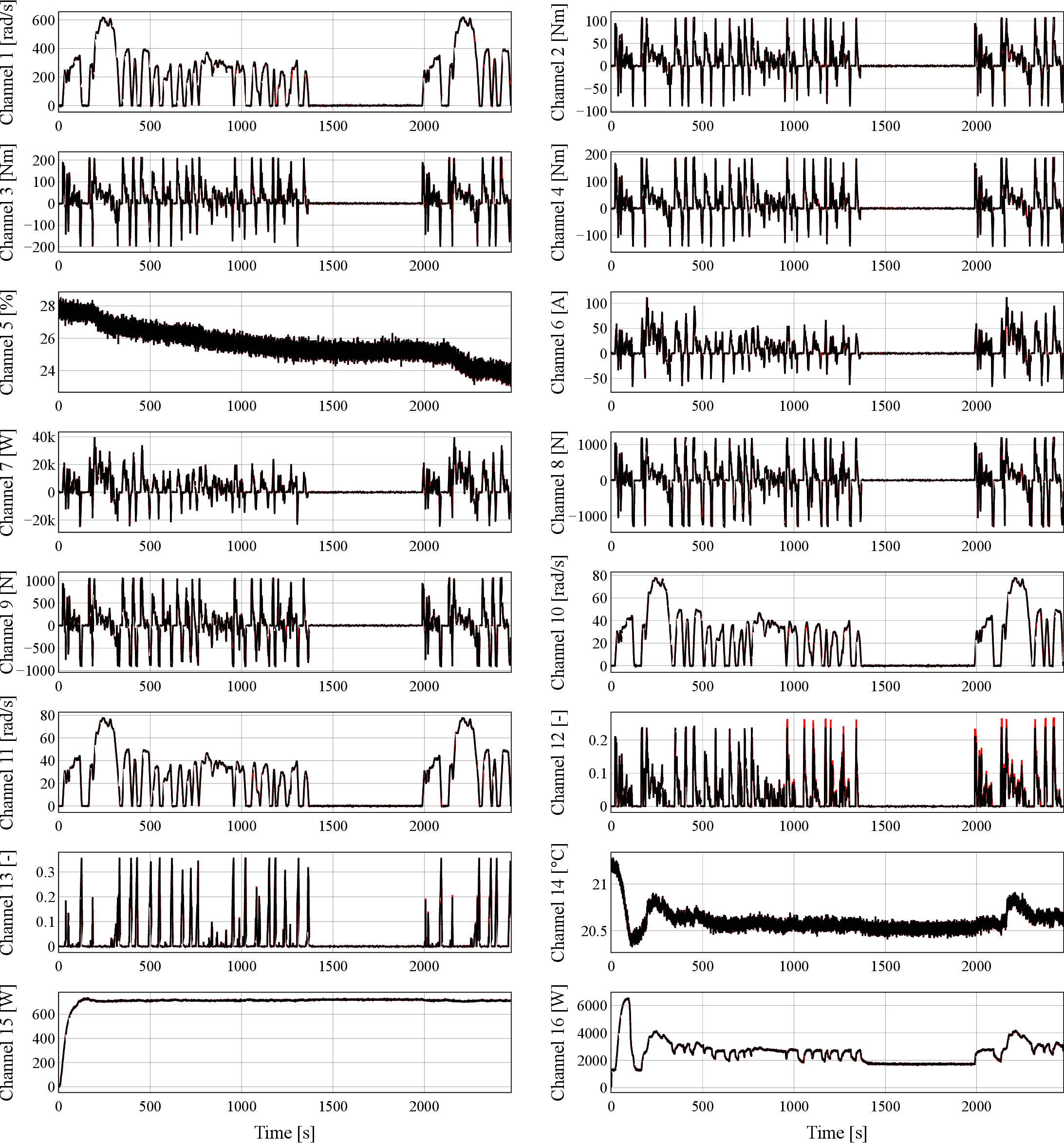}
    \caption{Plot of an \emph{anomalous} sequence with a reduced requested motor torque (in red) and its control counterpart (in black), both with added noise and undergone trimming.  The anomalous sub-sequence starts after \qty{940.2}{\second}. The channel legend can be found in Table~\ref{tab:signals}.}
    \label{fig:anomaly_plot_torquereq}
\end{figure}

For the next anomaly type, we reduce the requested motor torque value output by the powertrain control module by \qty{10}{\percent}. 
As a response to the change, the driver model requests a higher acceleration pedal value and consequently a different brake pedal values as well.
This anomaly type can also start from the beginning and from a random point in time within the cycle.
One of the anomalous time series for the FTP75 drive cycle and its control counterpart are plotted in Figure~\ref{fig:anomaly_plot_torquereq}. 

\begin{figure}[t!]
    \centering
    \includegraphics[width=0.98\textwidth]{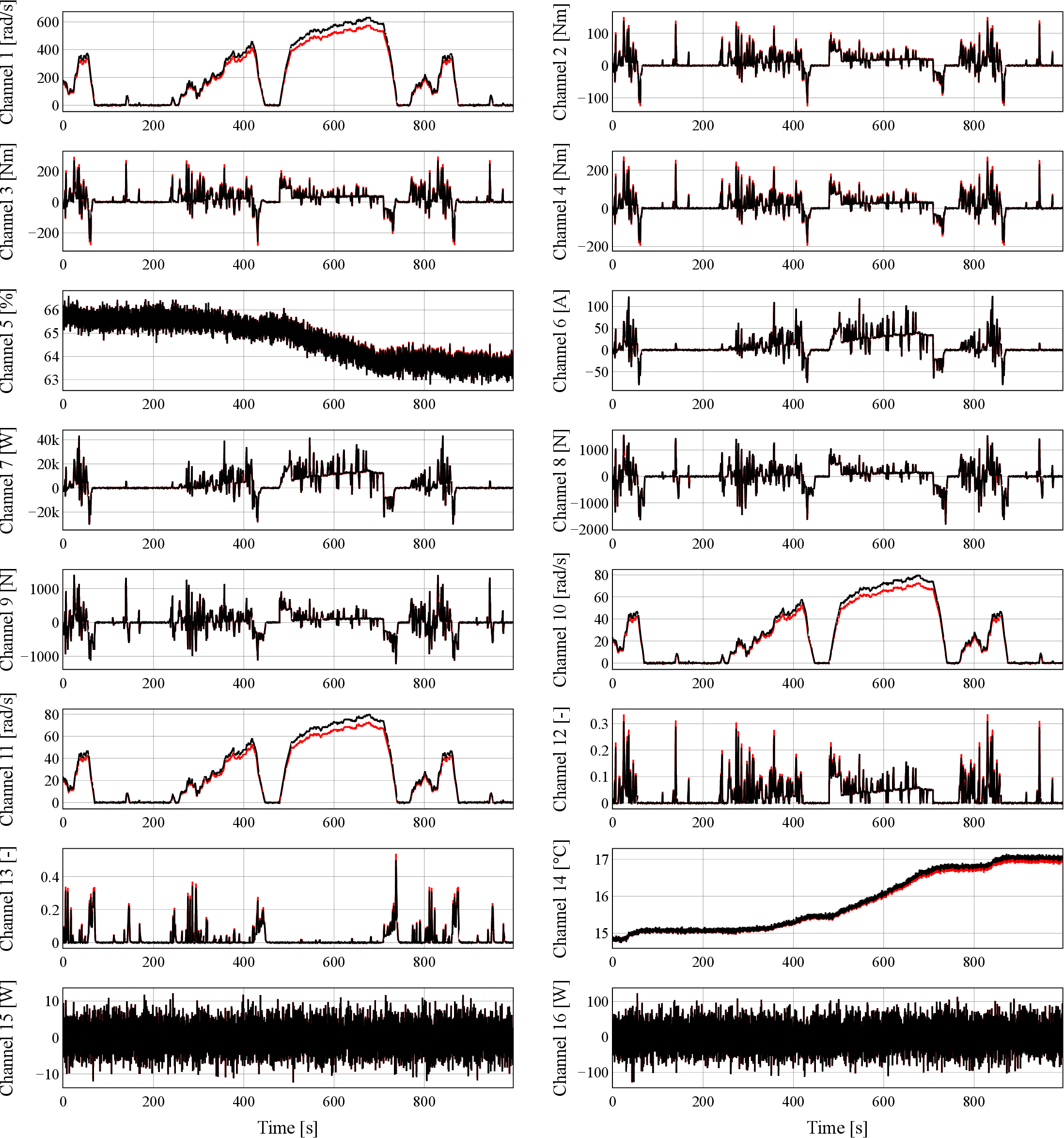}
    \caption{Plot of an \emph{anomalous} sequence with an increased loaded wheel diameter (in red) and its control counterpart (in black), both with added noise and undergone trimming. It is a whole-sequence anomaly, and hence the anomalous behaviour starts from the first time step. The channel legend can be found in Table~\ref{tab:signals}.}
    \label{fig:anomaly_plot_wheeldiameter}
\end{figure}

In the next case, we increase the loaded wheel diameter by \qty{10}{\percent} which, for the same target vehicle speed, leads to a lower motor and axle angular speed.
Furthermore, a larger wheel diameter leads to higher motor and axle torques, which are achieved using higher accelerator and brake pedal values.
Here, the wheel diameter also has an effect on the battery temperature, which, depending on its absolute magnitude, may also affect the cooling system.
Due to model limitations, this anomaly can only be simulated for whole-sequence anomalies.
One of the anomalous time series for the HUDDS drive cycle and its control counterpart are plotted in Figure~\ref{fig:anomaly_plot_wheeldiameter}. 

\begin{figure}[t!]
    \centering
    \includegraphics[width=0.98\textwidth]{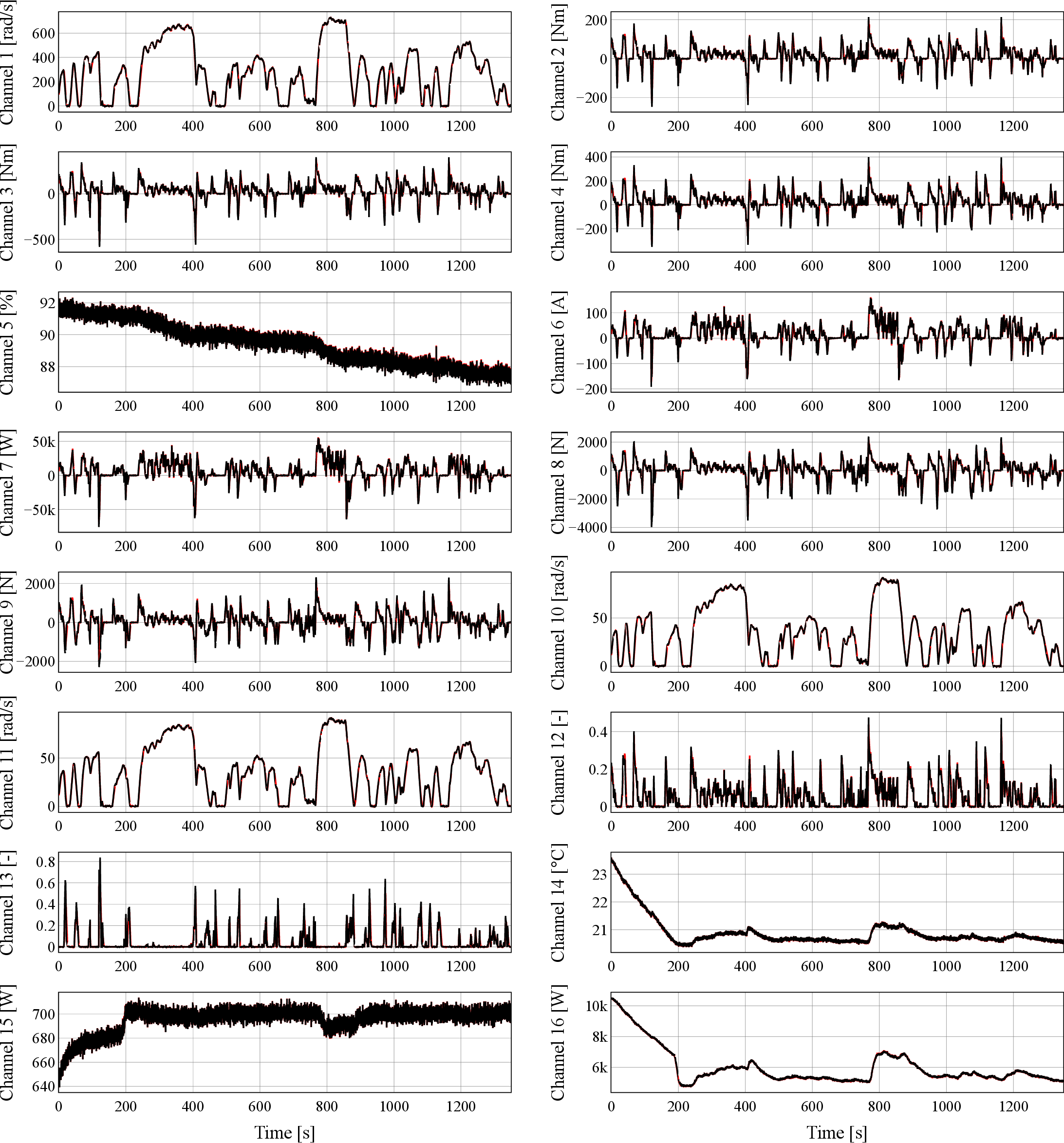}
    \caption{Plot of an \emph{anomalous} sequence with an increased driver response time (in red) and its control counterpart (in black), both with added noise and undergone trimming. It is a whole-sequence anomaly, and hence the anomalous behaviour starts from the first time step. The channel legend can be found in Table~\ref{tab:signals}.}
    \label{fig:anomaly_plot_driverreaction}
\end{figure}

The last anomaly is recorded after increasing the driver response time by a factor of $4$. 
This is one of the more subtle anomalies types, but manifests itself in all channels, except for the battery temperature and cooling.
Like for the wheel diameter anomaly, this anomaly can only be simulated for whole-sequence anomalies.
One of the anomalous time series for the LA92 drive cycle and its control counterpart are plotted in Figure~\ref{fig:anomaly_plot_driverreaction}. 

To ensure that the different anomaly types actually lead to anomalous behaviour, we run control simulations with the same initial battery states but with otherwise nominal model properties.
Given the uniform distribution from which the battery temperature is sampled from, half of the simulated anomaly types start with a battery temperature below \qty{20}{\degreeCelsius}.
In these cases, the battery will naturally heat up as it is being used and hence the cooling system does not play a role.
Therefore, in the case of the reduced cooling pump displacement, often no anomalous behaviour can be observed because the simulated anomaly is identical with the corresponding control simulation.
For these cases, the simulated anomaly is discarded.

Finally, this results in $N_\text{a}=284$ successful anomalous simulations, where $N_\text{a}=N_\text{ss}+N_\text{ws}$. Hence, the entire dataset $\mathcal{D}$ consists of $N_\text{n}+N_\text{ss}+N_\text{ws}=3557$ unique (nominal and anomalous) multivariate time series, with an anomalous sequence ratio of $284/3557 \approx 8\%$. 
$\mathcal{D}$ is then shuffled and divided into three separate folds for cross-validation, which corresponds to $2/3$ and $1/3$ split training and test subsets, respectively.
Formally, the training subset $\mathcal{D}^\text{train} = \{\mathcal{S}_1, ..., \mathcal{S}_m, ..., \mathcal{S}_{M}\}$ then consists of $M=2371$ multivariate time series on average, where $\mathcal{S}_m \in \mathbb{R}^{T_{m} \times d_\mathcal{D}}$, where $T_{m}$ is the number of time steps in sequence $\mathcal{S}_m$.
Likewise, the test subset $\mathcal{D}^\text{test} = \{\mathcal{S}_1, ..., \mathcal{S}_n, ..., \mathcal{S}_{N}\}$ consists of $N=1186$ multivariate time series on average, where $\mathcal{S}_n \in \mathbb{R}^{T_{n} \times d_\mathcal{D}}$, where $T_{n}$ is the number of time steps in sequence $\mathcal{S}_n$. For benchmarking purposes, we suggest the users use the prescribed training and test split to ensure comparable results.

To add further complexity and to reflect \lc{certain} real-world \lc{artefacts}, we undertake some \emph{post-processing}. First, we trim the beginning of each time series in $\mathcal{D}$ by random amounts so that time series representing the same drive cycle are rarely in sync.
The amount by which a given time series is trimmed is sampled from uniform distribution $\mathcal{U}(0, 0.1T)$.
This artefact can happen in the real world and means that, for the same drive cycle, any given time step is not comparable across different sequences, eliminating the viability of simple statistical methods such as control charts. 
In addition to that, we add noise sampled from Gaussian distribution $\mathcal{N}(0, 0.01\mathbf{\sigma}_\mathcal{D})$ to further move the obtained data towards the real world, where $\mathbf{\sigma}_\mathcal{D}$ is the feature-wise standard deviation of the dataset.

\subsection{Usability of the Dataset}
Clearly, both $\mathcal{D}^\text{train}$ and $\mathcal{D}^\text{test}$, as specified previously, contain nominal and anomalous time series, though in an unsupervised setting the labels for $\mathcal{D}^\text{train}$ should be disregarded.
This is because the dataset is aimed at \emph{unsupervised} time series anomaly detection, which requires approaches especially robust to contaminated training data.

We believe the underlying properties of the dataset can be useful in other research areas too.
The same $\mathcal{D}^\text{train}$ and $\mathcal{D}^\text{test}$ subsets can also be used for \emph{imbalanced time series classification} if the labels are considered.
Additionally, we provide a number of different subset variations for other tasks.
For \emph{semi-supervised} anomaly detection, we provide a clean $\mathcal{D}^\text{train}$ with on average $M=2182$ nominal time series and the same labelled $\mathcal{D}^\text{test}$ in the dataset, where clean refers to the absence of anomalous sequences in the subset.
Furthermore, for time series \emph{forecasting} or \emph{generation}, we supply clean versions of both $\mathcal{D}^\text{train}$ and $\mathcal{D}^\text{test}$, where $M=2182$ and $N=1091$ on average, respectively.
Despite being targeted at \emph{online} time series anomaly detection, the PATH dataset can just as well be used in offline time series anomaly detection.

\section{Baseline Results on the Dataset} \label{sec:results}
\subsection{Methodology}
\begin{figure}[t!]
    \centering
    \includegraphics[width=0.60\textwidth]{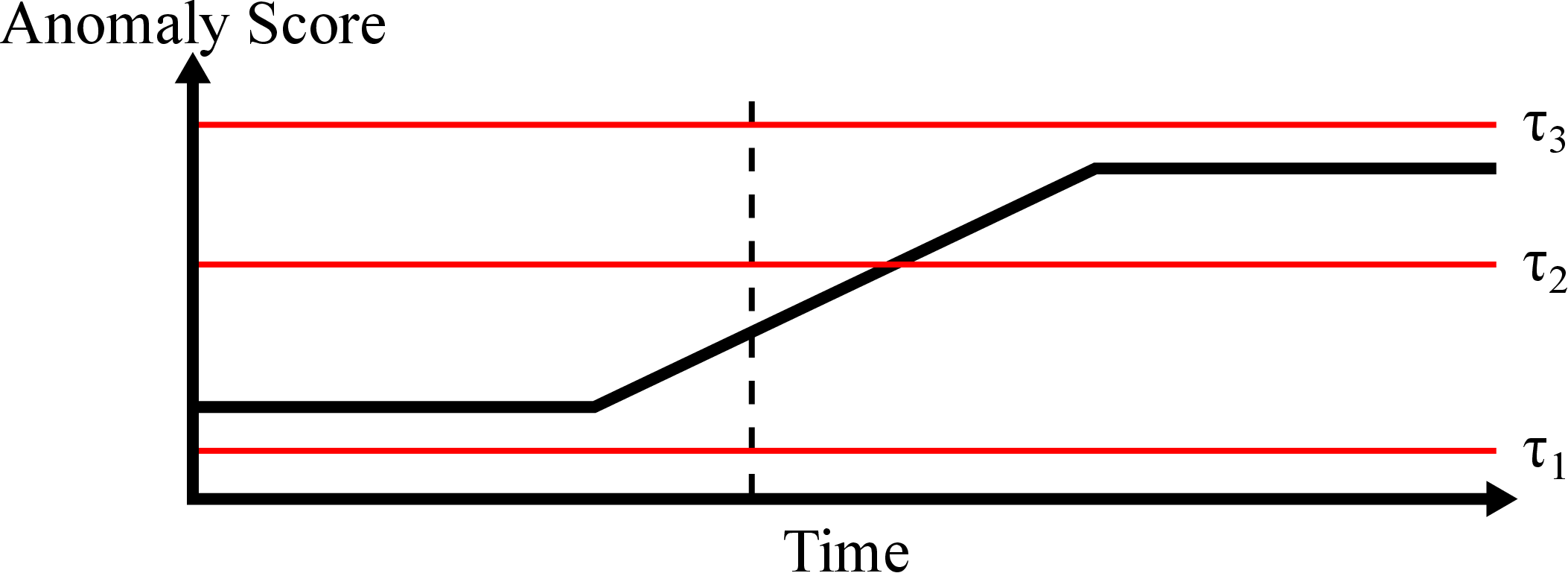}
    \caption{\lc{Arbitrary anomaly score as a function of time. The dotted vertical line denotes the start of the anomalous behaviour and the red lines represent three different thresholds.}}
    \label{fig:online_metrics}
\end{figure}

The evaluation metrics used to quantify anomaly detection performance by Correia et al.~\cite{correiaTeVAEVariationalAutoencoder2025} are adopted, as they provide a parameter-free way to quantify \emph{online} anomaly detection performance in an interpretable way.
Said metrics are very similar to the conventional true positive, false positive, true negative and false negative labels applied to each individual discrete sequence, with the exception that a sub-sequence anomaly can also be labelled as a false positive if detected too early.
In addition to that, the time between detection and ground-truth anomaly start is also quantified for each anomalous $\mathcal{S}_n$, with false positives being assigned the absolute value of the “negative” delay and false negatives being assigned the length of the anomalous sub-sequence within $\mathcal{S}_n$. 
The detection delays are finally aggregated into the average detection delay $\bar{\delta}$, given in seconds.
\lc{The issue with these metrics is that they are not \emph{recall consistent} as defined by Wagner et al.~\cite{wagnerTimeSeADBenchmarkingDeep2023}, meaning that the recall monotonically decreases with an increasing threshold.
Consider a time series with a sub-sequence anomaly starting off as nominal and eventually becoming anomalous, as shown in Figure~\ref{fig:online_metrics}.
For a very low threshold $\tau_1$, the anomaly is considered a false positive since it is an early detection.
As the threshold increases to $\tau_2$ it leads to a true positive but when the threshold reaches $\tau_3$, it becomes a false negative.
This leads to an increase and then decrease in recall, hence the metrics do not qualify as recall consistent, which also prevents them from being used to calculate the area under the precision-recall curve to quantify the uncallibrated detection performance.
Despite its short-comings, it is the only set of metrics that are apt for online discrete-sequence problems.}

In this work, we consider \lc{VS-VAE~\cite{pereiraUnsupervisedAnomalyDetection2018},} OmniAnomaly~\cite{su_robust_2019}, \lc{VASP~\cite{vonschleinitzVASPAutoencoderbasedApproach2021},} TCN-AE~\cite{thill_temporal_2021}, SISVAE~\cite{li_anomaly_2021}, LW-VAE~\cite{fahrmann_lightweight_2022}, TSADIS~\cite{tafazoli_matrix_2023}, and TeVAE~\cite{correiaTeVAEVariationalAutoencoder2025} when conducting experiments.
The hyperparameters for each approach are set as specified in the respective publication, though early stopping is applied to all that require a training procedure. 
Early stopping is parameterised such that the respective reconstruction error is monitored and training is stopped once validation loss has stopped decreasing for 250 epochs.

The validation subset $\mathcal{D}^\text{val}$ is obtained by further splitting $\mathcal{D}^\text{train}$ and hence is also unlabelled.
As future work may \lc{not} require a validation subset, it is left to the individual to extract it from the training subset if needed.
The test subset $\mathcal{D}^\text{test}$ should be the same as the one provided to ensure comparable results.

As mentioned in Section~\ref{sec:dataset} the simulation signals are sampled at \qty{10}{\hertz} by default, however, to reduce the computational load in our experiments, we downsample the data to \qty{2}{\hertz} with a low-pass filter with a cut-off frequency of \qty{1}{\hertz}, as it is consistent with the Whittaker–Nyquist–Shannon theorem~\cite{shannon_communication_1949}. 
This downsampling procedure is considered as part of the approaches tested and is optional for any future work, which may alternatively use the raw time series data or perhaps even correlation matrices~\cite{tayeh_attention-based_2022, zhang_deep_2019}.

To bring all channels to a common magnitude, the dataset features are z-score normalised.

Finally, we segment the time series data into fixed-length sub-sequences, also referred to as \emph{windows}. 
The rationale for using windows instead of full-length sequences is that the dynamics present in the time series data tend to occur quickly and only influence the data for a brief duration. 
Modelling entire variable-length sequences is possible, but it would lead to inefficient use of the model's learning capacity, as it would have to maintain information over unnecessarily long periods.
By focusing on windows that are just long enough to capture the existing dynamics in the data, model training should be more effective. 
To determine the optimal window length at \qty{2}{\hertz}, especially to capture the slowest dynamics present in a signal, we perform an autocorrelation analysis~\cite{correiaTeVAEVariationalAutoencoder2025} for each drive cycle and for every feature within those cycles, yielding a window size of $256$ time steps.
In the literature, the window size is often treated as a hyperparameter~\cite{fahrmann_lightweight_2022, tuli_tranad_2022, tafazoli_matrix_2023} or provided without reasoning~\cite{pereira_unsupervised_2018, chen_unsupervised_2020, li_anomaly_2021, thill_temporal_2021, doshi_reward_2022}. 
However, it is not possible to tune hyperparameters outside a supervised setting, and therefore such methods might not be applicable in real-world settings. 
In contrast, finding a suitable window length using autocorrelation is completely unsupervised. 
TSADIS takes window size as a hyperparameter before calling, hence a window size of $256$ time steps is also used.
To map the individual windows back to continuous sequences, mean-type reverse-windowing~\cite{correiaTeVAEVariationalAutoencoder2025} is used where applicable.

\subsection{Reproducibility and Benchmarking Considerations}
While perhaps sounding similar, repeatability, reproducibility, and replicability are defined differently according to the Association for Computing Machinery~\cite{ACM}.
\begin{itemize}
    \item Repeatability: the property of the research's finding being obtainable using the same experimental setup by the same person.
    \item Reproducibility: the property of the research's finding being obtainable using the same experimental setup by a \emph{different} person.
    \item Replicability: the property of the research's finding being obtainable using the \emph{different} experimental setup by a \emph{different} person.
\end{itemize}

Several position papers~\cite{drummond_replicability_2009, peng_reproducible_2011, munafo_manifesto_2017, gundersen_state_2018, bartz-beielstein_benchmarking_2020, kapoor_leakage_2023, semmelrock_reproducibility_2023} call for greater attention to be paid to reproducibility and replicability in computer science.
Additionally, some conferences focus on reproduction, like the  Machine Learning Reproducibility Challenge~\cite{MLRC}, or make specific calls for reproducibility and replicability papers, like the European Conference on Information Retrieval~\cite{ECIR}.
To enable future work to reproduce the results in this paper, we aim to be as transparent as possible by providing publicly available, clean and thoroughly commented source code for all experiments and the Simulink model under \url{https://github.com/lcs-crr/PATH}, as is suggested in literature~\cite{peng_reproducible_2011, munafo_manifesto_2017, semmelrock_reproducibility_2023}.

The seed for random operations has an impact on model training, given that processes like sampling and weight initialisation rely on it. 
To increase robustness of the results and to eliminate the possibility of the results owing to a specific fold and seed combination rather from the characteristics of the model~\cite{semmelrock_reproducibility_2023}, all three folds are trained on seeds $1$ through $3$, yielding $9$ different combinations.
The final result is then given as the average of the $9$ different combinations.
As mentioned, TSADIS does not require training, and hence its results are simply the average over all three folds. 

In case future work aims to replicate the results of this paper, we encourage deviating from the experimental setup outlined in this paper~\cite{goodman_what_2016}, though, as Bartz-Beielstein et al.~\cite{bartz-beielstein_benchmarking_2020} point out, there is no definition for how different an experimental setup needs to be for results to be considered replicable.
Using a different set of seeds, splitting the dataset into different folds, using different implementations of the approaches or even by using different software and hardware are some of the variables that could be changed in the setup, for example.
In the case of replicability, \lc{these} changes should not change the outcome~\cite{gundersen_state_2018}.
Moreover, it is just as important that future work provides the same level of transparency regarding the experimental setup and documentation.

It should be noted that the test subset $\mathcal{D}^\text{test}$ is often not available in the real world, so we strongly discourage approaches performing supervised threshold search using the labelled test data in $\mathcal{D}^\text{test}$.

Furthermore, there is no way to stop future research from performing hyperparameter tuning using $\mathcal{D}^\text{test}$, hence any results obtained for this dataset should be considered as the theoretical maximum anomaly detection performance achievable by the approach, not as a realistic anomaly detection performance observable in the real world.

We run all simulations that generate the PATH dataset on a workstation equipped with an Intel Xeon Gold 6234 CPU running Windows 10 Enterprise LTSC version 21H2 with MATLAB 2023b.
The framework used for model training is TensorFlow 2.15.1 and TensorFlow Probability 0.23 on Python 3.10 on a workstation running Ubuntu 22.04.5 LTS, equipped with two Nvidia RTX A6000 GPUs.
All work involving TSADIS is done in a separate environment with the latest compatible Python version of 3.9.
Further information on library versions used can be found in the \emph{requirements.txt} file in the repository.

\subsection{Results and Discussion}
To provide baseline results for the version of the PATH dataset for \emph{unsupervised} anomaly detection, we test several approaches. 
The corresponding results are shown in Table~\ref{tab:benchmark_us}.
\begin{table}[!t]
\caption{Mean $\pm$ standard deviation of $F_1$ score, precision $P$, recall $R$, and average detection delay $\bar{\delta}$ using the unsupervised threshold (top half) and theoretical best threshold (bottom half) for a range of approaches applied to the \emph{unsupervised} anomaly detection version of the PATH dataset, i.e.\ the version with anomalous data in training subset $\mathcal{D}^\text{train}$. The best $F_1$ scores are given in \textbf{bold}.}\label{tab:benchmark_us}
\centering
\setlength{\tabcolsep}{3pt}
\begin{tabular}{lcccccc}
Approach & $F_1$            & $P$              & $R$              & $\bar{\delta} \ [s]$ \\ \hline\hline
VS-VAE  & $0.03 \pm 0.02$  & $0.78 \pm 0.41$  & $0.01 \pm 0.01$  & $\phantom{0}991.4 \pm 71.1\phantom{x}$   \\ 
OmniA   & $\mathbf{0.06 \pm 0.06}$  & $0.60 \pm 0.44$  & $0.03 \pm 0.03$  & $\phantom{0}994.9 \pm 69.7\phantom{x}$   \\ 
VASP    & $0.02 \pm 0.02$  & $0.21 \pm 0.34$  & $0.01 \pm 0.01$  & $\phantom{0}991.9 \pm 70.8\phantom{x}$   \\ 
TCN-AE  & $0.02 \pm 0.02$  & $0.45 \pm 0.41$  & $0.01 \pm 0.01$  & $\phantom{0}989.8 \pm 66.9\phantom{0}$   \\ 
SISVAE  & $0.03 \pm 0.03$  & $0.13 \pm 0.10$  & $0.02 \pm 0.02$  & $\phantom{0}993.1 \pm 69.6\phantom{0}$   \\ 
LW-VAE  & $0.01 \pm 0.02$  & $0.26 \pm 0.41$  & $0.01 \pm 0.01$  & $\phantom{0}991.9 \pm 71.0\phantom{0}$   \\ 
TSADIS  & $0.00 \pm 0.00$  & $0.00 \pm 0.00$  & $0.00 \pm 0.00$  & $1209.8 \pm 8.9\phantom{00}$   \\ 
TeVAE   & $0.02 \pm 0.03$  & $0.27 \pm 0.36$  & $0.01 \pm 0.02$  & $\phantom{0}992.7 \pm 70.7\phantom{0}$   \\ \hline
VS-VAE  & $0.30 \pm 0.07$  & $0.35 \pm 0.26$  & $0.34 \pm 0.05$  & $\phantom{0}792.0 \pm 75.4\phantom{x}$   \\ 
OmniA   & $0.50 \pm 0.09$  & $0.60 \pm 0.11$  & $0.44 \pm 0.11$  & $\phantom{0}767.0 \pm 114.7$   \\ 
VASP    & $0.11 \pm 0.01$  & $0.07 \pm 0.01$  & $0.47 \pm 0.24$  & $\phantom{0}673.5 \pm 181.2$   \\ 
TCN-AE  & $0.14 \pm 0.01$  & $0.09 \pm 0.01$  & $0.41 \pm 0.16$  & $\phantom{0}750.2 \pm 121.1$   \\ 
SISVAE  & $0.22 \pm 0.03$  & $0.19 \pm 0.05$  & $0.28 \pm 0.04$  & $\phantom{0}826.2 \pm 80.0\phantom{0}$   \\
LW-VAE  & $0.13 \pm 0.01$  & $0.07 \pm 0.01$  & $0.51 \pm 0.22$  & $\phantom{0}634.8 \pm 172.6$   \\ 
TSADIS  & $0.10 \pm 0.01$  & $0.05 \pm 0.00$  & $1.00 \pm 0.00$  & $1209.8 \pm 8.9\phantom{00}$   \\ 
TeVAE   & $\mathbf{0.58 \pm 0.08}$  & $0.69 \pm 0.12$  & $0.50 \pm 0.09$  & $\phantom{0}676.4 \pm 103.2$   \\ \hline
\end{tabular}
\end{table}

First, it is evident that there is a large gap between the unsupervised and theoretical best threshold results. 
The unsupervised threshold is a rudimentary estimation based on the unlabelled validation subset $\mathcal{D}^\text{val}$~\cite{correiaTeVAEVariationalAutoencoder2025} and tends to be set higher than the theoretical best.
This is because in the unsupervised version of the dataset, there are anomalous sequences within $\mathcal{D}^\text{val}$, which are associated with a higher maximum anomaly score and therefore threshold.
It is clear, however, that the results are far from good, which sets one foundation for future work.
\lc{We can isolate absolute detection performance from} the threshold choice by performing a grid search on different thresholds, which allows us to find the threshold yielding the theoretical best $F_1$ score.
At this threshold, we observe the best possible anomaly detection performance the approach can achieve on the test set, though it is not observable in the real-world.
Here, TeVAE performs best in terms of $F_1$ score, though with a high average detection delay due to the number of high number of false negatives.
While these results are better than with the unsupervised threshold, they still leave room for improvement.
Regardless of the results, it cannot be denied how much less computationally intensive TSADIS is compared to methods based on deep learning.
Even on commodity hardware, more specifically a laptop with an Intel Core i7-1185G7, it can evaluate test data faster than deep learning models. 
It should be noted that this is mainly because no reverse-windowing is needed with TSADIS, a process that, unlike inference in deep learning models, runs on the CPU, not the GPU.
\lc{However, because TSADIS works on the entire sequence, not on windows, it is technically an offline approach, hence why its detection delay is the highest.}
Additionally, TSADIS does \emph{not} require a training procedure. At first glance, this property is a benefit, as the implementation hurdle is much lower than with deep learning models, which essentially require GPU-acceleration. 
However, without training data, TSADIS cannot know what is a nominal time series and what is not, therefore the nominal behaviour is \emph{not modelled}.
It can solely rely on the information present within a sequence to calculate an anomaly score, which is part of the reason it cannot outperform deep learning-based methods.

We also performed limited testing on the version of the dataset for \emph{semi-supervised} anomaly detection. 
It involves the same testing procedure, except that the clean version, i.e.\ anomaly-free, of the training subset is used.

The corresponding results for TeVAE are shown in Table~\ref{tab:benchmark_ss}.
\begin{table}[!t]
\caption{Mean $\pm$ standard deviation of $F_1$ score, precision $P$, recall $R$, and average detection delay $\bar{\delta}$ using the unsupervised threshold (top half) and theoretical best threshold (bottom half) for TeVAE applied to the \emph{semi-supervised} anomaly detection version of the PATH dataset, i.e.\ the version without anomalous data in training subset $\mathcal{D}^\text{train}$.}\label{tab:benchmark_ss}
\centering
\setlength{\tabcolsep}{3pt}
\begin{tabular}{lcccccc}
Approach& $F_1$            & $P$              & $R$              & $\bar{\delta} \ [s]$ \\ \hline\hline
TeVAE   & $0.63 \pm 0.22$  & $0.95 \pm 0.06$  & $0.52 \pm 0.26$  & $\phantom{0}589.1 \pm 212.0$   \\ \hline
TeVAE   & $0.80 \pm 0.15$  & $0.88 \pm 0.11$  & $0.76 \pm 0.18$  & $\phantom{0}412.6 \pm 143.8$   \\ \hline
\end{tabular}
\end{table}
The gap between results obtained using the unsupervised threshold and the theoretical best is now much smaller than observed in Table~\ref{tab:benchmark_us}, which can be attributed to the lack of anomalous data in $\mathcal{D}^\text{val}$.
Additionally, there is a large gap between the theoretical best results between the unsupervised and semi-supervised versions, indicating the need for more robust future work when anomalous data is present in $\mathcal{D}^\text{train}$.

\section{Conclusion and Outlook} \label{sec:conclusion}
We propose a novel multivariate time series dataset for online anomaly detection, called the Powertrain Anomaly Time series bencHmark (PATH) dataset.
The PATH dataset is generated using simulation, where the model it is based on resembles a real-world dynamic system. 
In addition to that, simulation is done in a variety of different initial states to further add to the diversity of the dataset.
To increase the complexity of the dataset, noise is applied and the beginning of time series are randomly trimmed.
The anomalies in the PATH dataset arise from changing parameters prior to simulation, as opposed to manual data manipulation, resulting in anomalies that are non-trivial and realistic.
We offer the dataset in three different versions: one for unsupervised anomaly detection, where the training subset consists of both anomalous and nominal sequences, another for semi-supervised anomaly detection, where the training subset consists of nominal sequences only, and one for time series generation or forecasting, where both the training and test subsets are nominal.
Lastly, for each of the versions, we offer three different folds with a pre-defined train and test split to ensure generalised and comparable results. 

The experiments conducted in this work further support the claim of non-triviality because, even when the threshold choice is removed as a factor, the best approach in an unsupervised setting only manages to achieve an $F_1$ score of $0.58$ and an average detection delay of \qty{676.4}{\second}.
In contrast, however, the results significantly improve when the clean version of the test subset is used.
Here, the average theoretical best $F_1$ score reaches $0.80$ and an average detection delay of \qty{412.6}{\second}, highlighting the need for methods more robust to anomalous data in the unlabelled and contaminated training subset.

In the future, the PATH dataset should be extended to a standardised benchmark consisting of not only a dataset based on the longitudinal electric vehicle dynamics, but also on simulation models from other domains.
Additionally, the simulation model can be adapted to take \lc{other factors into account}, like battery ageing. 
Battery ageing can be characterised by the charge capacity, which, fixed in this dataset, can be changed dynamically to simulate battery ageing, which will have an impact on the entire system.
This property can be especially useful for research in the area of unsupervised predictive maintenance, where an explicit health signal is not present.
\lc{Additionally, modal channels could be considered if a gearbox model was implemented, since, depending on the discrete gear, the motor speed is decoupled from the axle speed.}
Furthermore, there is a need for more sophisticated evolution of the online evaluation metrics~\cite{correiaTeVAEVariationalAutoencoder2025} that do not assume a \lc{single} anomalous sub-sequence per test time series \lc{and that are recall consistent}.
When said metrics are available, the dataset could be extended to such anomalies types.

\printbibliography 

\end{document}